\documentclass[journal]{IEEEtran}
\usepackage[T1]{fontenc}
\usepackage{cite}
\usepackage{amsmath,amssymb,amsfonts}
\usepackage{graphicx}
\usepackage{booktabs}
\usepackage{multirow}
\usepackage{algorithm}
\usepackage{algorithmic}
\usepackage{tikz}
\usetikzlibrary{shapes.geometric,arrows.meta,positioning,fit}
\usepackage[hidelinks]{hyperref}

\graphicspath{{figures/}}

\newcommand{\bdix}{BDIx}
\newcommand{\ket}[1]{\left|#1\right\rangle}

\begin{document}

\title{A Quantum-Assisted Agentic Distributed Artificial Intelligence Framework for Deadline-Bounded Orchestration of Hybrid Renewable Microgrids}

\author{Iacovos~I.~Ioannou,~\IEEEmembership{Senior~Member,~IEEE,}
        Saher~Javaid,~\IEEEmembership{Member,~IEEE,}
        Minella~Bezha,\\
        Yasuo~Tan,~\IEEEmembership{Member,~IEEE,}
        Naoto~Nagaoka
        and~Vasos~Vassiliou,~\IEEEmembership{Senior~Member,~IEEE}%
\thanks{I. I. Ioannou is with the Department of Computer Science, Philips University, Nicosia, Cyprus and also with the Department of Computer Science, University of Cyprus and the CYENS Centre of Excellence, Nicosia, Cyprus (e-mail: ioannou.iacovos@ucy.ac.cy).}%
\thanks{S. Javaid and Y. Tan are with the School of Information Science, Japan Advanced Institute of Science and Technology, Nomi, Ishikawa, Japan.}%
\thanks{M. Bezha and N. Nagaoka  are with the International Infrastructure System Research Center, Doshisha University, Kyoto, Japan.}%
\thanks{V. Vassiliou is with the Department of Computer Science, University of Cyprus and the CYENS Centre of Excellence, Nicosia, Cyprus.}}

\markboth{IEEE Transactions on Smart Grid,~Vol.~XX, No.~X, 2026}%
{Ioannou \MakeLowercase{\textit{et al.}}: A Quantum-Assisted Agentic DAI Framework for Deadline-Bounded Microgrid Orchestration}

\maketitle

\begin{abstract}
The real-time orchestration of microgrids that combine fluctuating renewable sources, constant dispatchable units, storage and curtailable consumers requires the repeated solution of combinatorial dispatch and coalition formation problems under hard control deadlines. In this paper, a quantum-assisted agentic distributed artificial intelligence (DAI) framework is proposed in which the dispatch problem of each control slot is formulated as a quadratic unconstrained binary optimization (QUBO) problem by Belief-Desire-Intention extended (\bdix{}) agents operating within the framework and is solved by a portfolio of quantum, quantum-inspired and classical solvers. Solver selection is elevated to a first-class agentic deliberation action of the coordinator agent. Learned beliefs about solver latencies are maintained and the solver intention that is expected to satisfy the prevailing deliberation deadline is committed in each slot. In addition, a belief-shaped storage valuation mechanism is introduced through which the storage agent prices its energy at a discounted future-peak value, thereby injecting intertemporal information into the otherwise myopic per-slot optimization. The framework is evaluated on a 24-hour simulation of a grid-connected microgrid comprising photovoltaic, wind, battery, genset and demand-response assets, with the Quantum Approximate Optimization Algorithm (QAOA) executed by statevector simulation and benchmarked per slot against tabu search, simulated annealing, binary particle swarm optimization, greedy descent and exhaustive enumeration. The results show that zero deliberation deadlines are missed, that the committed dispatch attains the exact optimum on every slot and that the realized daily cost equals the exact lower bound. A daily operating cost of 146.24 EUR, renewable utilization of 97.83 percent, zero unserved energy and zero deliberation deadline misses are recorded in the final trace. When the storage valuation mechanism is deactivated, the daily operating cost is increased to 152.75 EUR, corresponding to a 4.5 percent increase.
\end{abstract}


\IEEEpeerreviewmaketitle

\section{Introduction}
\IEEEPARstart{T}{he} proliferation of distributed energy resources has transformed the distribution edge into a collection of mini power grids in which photovoltaic (PV) plants, wind turbines, battery energy storage systems (BESS), dispatchable gensets and flexible consumers must be coordinated in real time \cite{lasseter2002,olivares2014,zia2018}. The energy management system (EMS) of such a microgrid is required to commit, at every control slot, a set of discrete decisions that includes the commitment of the genset, the charging or discharging of the storage, the curtailment of flexible loads and the activation of grid exchange blocks. The resulting per-slot problem is a combinatorial optimization problem whose size grows exponentially with the number of controllable assets and which must nevertheless be solved within deadlines that range from hundreds of milliseconds during routine operation down to tens of milliseconds during critical transients such as islanding events \cite{olivares2014}.

Two research directions have been pursued in order to address this tension. On the one hand, multi-agent system (MAS) architectures distribute the sensing, reasoning and actuation of the EMS across software agents that are attached to the physical assets \cite{dimeas2005,logenthiran2012,mcarthur2007}. On the other hand, quantum and quantum-inspired optimization has been investigated as an accelerator for the combinatorial subproblems of power system operation, with unit commitment, optimal power flow and facility placement formulated as quadratic unconstrained binary optimization (QUBO) problems and solved by quantum annealers or by the Quantum Approximate Optimization Algorithm (QAOA) \cite{ajagekar2019,eskandarpour2020,morstyn2023,farhi2014,lucas2014}. These two directions have, however, evolved largely in isolation. Existing quantum-assisted dispatch studies treat the quantum solver as an external oracle that is invoked unconditionally, whereas existing MAS-based EMS designs employ classical heuristics whose runtime characteristics are not reasoned about explicitly. In the noisy intermediate-scale quantum (NISQ) era \cite{preskill2018}, in which quantum solver latencies are both significant and stochastic, the decision of \emph{when} a quantum solver should be invoked is as important as the decision of \emph{what} it should solve.

In this paper, the above gap is addressed by integrating quantum optimization into the deliberation cycle of the distributed artificial intelligence (DAI) framework, under which Belief-Desire-Intention extended (\bdix{}) agents operate \cite{rao1995,ioannou2020,ioannou2022cn,ioannou2022access}. The \bdix{} agent constitutes the agent model through which the DAI framework is realized. The classical BDI architecture \cite{bratman1987,rao1995} is extended with belief revision enhanced by machine learning. The model has previously been applied by the authors to device-to-device communication orchestration in 5G and beyond networks \cite{ioannou2020,ioannou2022cn,ioannou2022access}. It has also been consolidated in a dedicated book on the DAI framework for 5G and 6G communications \cite{daibook} and has been applied in the energy domain to nano-grid and mini-grid power management and control \cite{ioannou2024access}. In the present paper this line of work is extended in two directions. First, the invocation of a particular optimization solver is modeled as a first-class intention of the coordinator agent. The coordinator maintains learned beliefs about the latency of every solver in its deliberation portfolio, which comprises a statevector-simulated QAOA solver, a quantum-inspired tabu search solver and a greedy fallback. The solver intention that is expected to meet the prevailing deliberation deadline is committed in each slot. Second, a belief-shaped storage valuation mechanism is introduced through which the storage agent prices the energy in its battery at a discounted future-peak value derived from the tariff forecast. The effective charge and discharge prices that result from this valuation are embedded into the per-slot QUBO and inject intertemporal information into an optimization that would otherwise be myopic.

The contributions of this paper are summarized as follows.
\begin{enumerate}
\item A quantum-assisted DAI architecture for microgrid orchestration is proposed and realized by \bdix{} agents, in which the per-slot dispatch and coalition formation problem is assembled as a QUBO from distributed agent bids and quantum, quantum-inspired and classical solvers constitute alternative deliberation actions of the coordinator.
\item An agentic deadline-bounded meta-deliberation mechanism is introduced in which solver latencies are learned online as beliefs and the solver intention is selected per slot so that hard deliberation deadlines, including tightened deadlines during critical transients, are respected.
\item A belief-shaped storage valuation mechanism is introduced through which the future value of stored energy is injected into the myopic per-slot QUBO via effective charge and discharge prices computed by the storage agent.
\item A fully reproducible evaluation of the framework over a 24-hour scenario is provided, in which QAOA is benchmarked per slot against tabu search, simulated annealing, binary particle swarm optimization (PSO), greedy descent and exhaustive enumeration on identical QUBO instances with exact ground truth and in which an ablation study quantifies the contribution of each mechanism.
\end{enumerate}

The proposed framework is positioned as ready for quantum hardware deployment. The QAOA solver is executed in this paper by exact statevector simulation, which permits a controlled and reproducible evaluation. The solver interface is preserved so that the simulated backend can be replaced by gate-based hardware or by a quantum annealer without any modification to the agent layer.

The remainder of the paper is organized as follows. Related work and background information are presented in Section~\ref{sec:related}. The system model and the DAI architecture with its \bdix{} agents are presented in Section~\ref{sec:model}. The methodology is presented in Section~\ref{sec:method}, in which the QUBO formulation is derived in Section~\ref{sec:qubo}, the deadline-bounded quantum-classical deliberation mechanism is described in Section~\ref{sec:deliberation}, the belief-shaped storage valuation is described in Section~\ref{sec:storage} and the overall execution is summarized in Section~\ref{sec:overall}. The performance evaluation is reported in Section~\ref{sec:eval} and the paper is concluded in Section~\ref{sec:conclusion}.

\section{Related Work and Background Information}
\label{sec:related}
A review of the literature related to the proposed framework and the necessary background information are provided in this section. The related work is organized into multi-agent energy management and quantum optimization in power systems. The contribution of this paper is positioned against each branch, while the background information covers the DAI framework and the \bdix{} agents through which it is realized.

\subsection{Multi-Agent Energy Management}
MAS architectures for microgrid control were established by early works on agent-based market participation and islanded operation \cite{dimeas2005,mcarthur2007} and were subsequently applied to real-time operation with distributed intelligence located at generation, storage and load assets \cite{logenthiran2012}. Comprehensive reviews of microgrid EMS techniques \cite{zia2018,olivares2014} identify centralized mathematical programming, heuristic search and agent-based decomposition as the dominant solution families. Demand-side flexibility has been integrated through demand response programs whose discomfort costs are traded against generation costs \cite{palensky2011}. Game-theoretic formulations, including coalitional games among prosumers and storage owners, have also been employed for the smart grid \cite{saad2012}. At the algorithmic level, the centralized EMS formulations considered in this review are represented by two families. Metaheuristic dispatch is represented by the particle swarm optimization (PSO) based energy and operation management of Radosavljevi\'{c} \emph{et al.} \cite{radosavljevic2016}, in which the nonlinear economic operation problem is solved without convexity assumptions. Learning-based operation is represented by the deep reinforcement learning (DRL) scheme of Du and Li \cite{du2020}, in which trained policies deliver fast inference for multi-microgrid management at the expense of training complexity and limited interpretability. These formulations are optimized centrally and no formulation reasons about its own computational latency against control deadlines. The framework proposed in this paper belongs to the agent-based family but differs from prior MAS designs in that the reasoning of the coordinator covers not only the dispatch decision itself but also the computational process by which the decision is obtained.

\subsection{Quantum Optimization in Power Systems}
The application of quantum computing to energy system optimization has been surveyed in \cite{ajagekar2019,eskandarpour2020}. Unit commitment and optimal power flow have been mapped to Ising and QUBO formulations \cite{lucas2014,glover2019} and solved by annealing-based devices \cite{morstyn2023}, while variational algorithms such as QAOA \cite{farhi2014,zhou2020,cerezo2021} have been studied for combinatorial subproblems at NISQ scale \cite{preskill2018}. In the microgrid context specifically, the unit commitment problem was decomposed in \cite{nikmehr2022} through a quantum variant of the alternating direction method of multipliers and the resulting subproblems were assigned to quantum oracles, while QAOA was adapted to the unit commitment problem in \cite{koretsky2021} and combinatorial optimal power flow was solved by quantum annealing in \cite{morstyn2023}. These studies demonstrate feasibility but typically invoke the quantum solver unconditionally and offline. Control deadlines are not considered and the solver invocation is not embedded into an agentic decision process. In contrast, in this paper the quantum solver is one of several deliberation actions whose stochastic latency is reasoned about explicitly by the coordinator agent.

\subsection{Background Information: The DAI Framework and \bdix{} Agents}
The BDI model of agency \cite{bratman1987,rao1995} structures rational action around beliefs, desires and intentions and underlies a large body of agent programming research. The \bdix{} agent was introduced by the authors as the agent model of their DAI framework. BDI agents are augmented with belief revision assisted by machine learning and with distributed plan libraries. The model has been validated for the orchestration of device-to-device communication in cellular networks \cite{ioannou2020,ioannou2022cn,ioannou2022access}, which constitutes a key enabler of cooperative wireless architectures \cite{liu2009coop}, while a comprehensive treatment of the framework, its architecture and its machine learning plan composition is provided in \cite{daibook}. The plan library of the \bdix{} agent, through which beliefs are used to prioritize desires into intentions, was enhanced in \cite{ioannou2025planlib} through prioritization based on machine learning and fuzzy logic for 6G device-to-device communications. Closest to the present work, the DAI framework with \bdix{} agents was applied in \cite{ioannou2024access} to nano-grid and mini-grid power management and control, in which every power source, load and storage device is governed by a \bdix{} agent and the power balancing plans are realized by classical techniques including linear programming, genetic algorithms, ant colony optimization and PSO. The present paper extends that line of work by formulating the per-slot dispatch as a QUBO, by introducing quantum and quantum-inspired solvers into the plan library and by elevating the deadline-bounded selection among them to a deliberation action of the coordinator. To the best of the authors' knowledge, this is the first work in which quantum optimization solvers are modeled as \bdix{} deliberation actions within the DAI framework.

\section{System Model and DAI-Based \bdix{} Agent Architecture}
\label{sec:model}
The model of the considered microgrid is described in this section together with the DAI-based \bdix{} agent architecture through which the microgrid is operated. The reasoning cycle that is executed at every control slot is also described.

\subsection{Microgrid Model}
A grid-connected microgrid with islanding capability is considered over a horizon of $T$ control slots of duration $\Delta t$. In the evaluated scenario $T=96$ and $\Delta t = 0.25$~h, corresponding to one day at a 15-minute EMS resolution. The microgrid comprises the following assets, each of which is governed by a dedicated \bdix{} agent.
\begin{itemize}
\item A PV plant of capacity $P^{\mathrm{pv}}$ whose output $R^{\mathrm{pv}}_t$ follows a solar irradiance profile with smoothed cloud transients (fluctuating source).
\item A wind turbine of capacity $P^{\mathrm{w}}$ whose output $R^{\mathrm{w}}_t$ is a smoothed stochastic process (fluctuating source).
\item An inflexible aggregate load with demand $D^{\mathrm{b}}_t$ exhibiting morning and evening peaks (fluctuating consumer).
\item $K$ curtailable flexible loads with constant demands $P^{\mathrm{f}}_k$ and discomfort prices $c^{\mathrm{f}}_k$ in EUR/kWh, $k=1,\dots,K$ (constant flexible consumers).
\item A BESS with power rating $P^{\mathrm{s}}$, energy capacity $E^{\mathrm{s}}$, charge and discharge efficiencies $\eta_{\mathrm{c}}$ and $\eta_{\mathrm{d}}$, degradation price $c_{\mathrm{deg}}$ and state of charge (SoC) $s_t \in [s_{\min}, s_{\max}]$.
\item A diesel genset which, when committed, produces $P^{\mathrm{g}}$ at fuel price $c^{\mathrm{g}}$ (constant dispatchable source).
\item A point of common coupling (PCC) through which power is imported in $n_I$ binary blocks of sizes $P^{\mathrm{imp}}_j$ at the time-of-use tariff $\pi_t$ and exported in $n_E$ binary blocks of sizes $P^{\mathrm{exp}}_j$ at the feed-in tariff $\pi^{\mathrm{exp}}$. Residual deficits below the block granularity are settled by a continuous regulation slack of capacity $P^{\mathrm{slk}}$ priced at $\kappa \pi_t$ with $\kappa > 1$ and any remaining deficit constitutes unserved energy priced at the value of lost load $c_{\mathrm{VoLL}}$.
\end{itemize}
The renewable infeed and the total demand of slot $t$ are denoted by $R_t = R^{\mathrm{pv}}_t + R^{\mathrm{w}}_t$ and $D_t = D^{\mathrm{b}}_t + \sum_{k=1}^{K} P^{\mathrm{f}}_k$ respectively.

Each slot carries a deliberation deadline $\Delta^{\mathrm{dl}}_t$ within which the dispatch decision must be committed. Routine slots carry the deadline $\Delta^{\mathrm{dl}}_{\mathrm{n}}$. A fraction $p_{\mathrm{c}}$ of the slots is flagged as critical. These slots represent transients such as islanding or frequency events and carry the tightened deadline $\Delta^{\mathrm{dl}}_{\mathrm{c}} \ll \Delta^{\mathrm{dl}}_{\mathrm{n}}$.

\subsection{\bdix{} Reasoning Cycle}
\begin{figure}[!t]
\centering
\begin{tikzpicture}[
  font=\scriptsize,
  agent/.style={draw,rounded corners=1.5pt,minimum width=1.12cm,minimum height=0.62cm,align=center,fill=gray!8},
  coord/.style={draw,rounded corners=2pt,minimum width=6.6cm,minimum height=1.55cm,align=center,fill=gray!15},
  env/.style={draw,rounded corners=2pt,minimum width=6.6cm,minimum height=0.55cm,align=center,fill=gray!4},
  arr/.style={-{Stealth[length=2mm]},thick}
]
\node[agent] (pv)  {PV\\agent};
\node[agent,right=0.12cm of pv] (wt)  {Wind\\agent};
\node[agent,right=0.12cm of wt] (ld)  {Load\\agent};
\node[agent,right=0.12cm of ld] (fl)  {Flex load\\agents};
\node[agent,right=0.12cm of fl] (bs)  {Storage\\agent};
\node[agent,right=0.12cm of bs] (gn)  {Genset\\agent};
\node[coord,below=0.85cm of ld.south east,anchor=north] (co) {%
\textbf{Coordinator \bdix{} agent}\\[1pt]
beliefs: tariff, deadline, solver latencies $\hat{\ell}_\sigma$\\
deliberation: QUBO assembly $\rightarrow$ solver intention\\
portfolio: QAOA $\mid$ tabu $\mid$ greedy};
\node[env,below=0.5cm of co] (envb) {Microgrid environment (profiles, tariffs, PCC, SoC dynamics)};
\draw[arr] (pv.south) -- ++(0,-0.32) -| ([xshift=-2.7cm]co.north);
\draw[arr] (wt.south) -- ++(0,-0.32) -| ([xshift=-1.65cm]co.north);
\draw[arr] (ld.south) -- ([xshift=-0.6cm]co.north);
\draw[arr] (fl.south) -- ([xshift=0.6cm]co.north);
\draw[arr] (bs.south) -- ++(0,-0.32) -| ([xshift=1.65cm]co.north);
\draw[arr] (gn.south) -- ++(0,-0.32) -| ([xshift=2.7cm]co.north);
\draw[arr] (co.south) -- node[right]{setpoints $x_t$} (envb.north);
\draw[arr] (envb.west) -- ++(-0.32,0) |- node[pos=0.25,left]{perception} (pv.west);
\end{tikzpicture}
\caption{Quantum-assisted DAI architecture realized by \bdix{} agents.}
\label{fig:arch}
\end{figure}
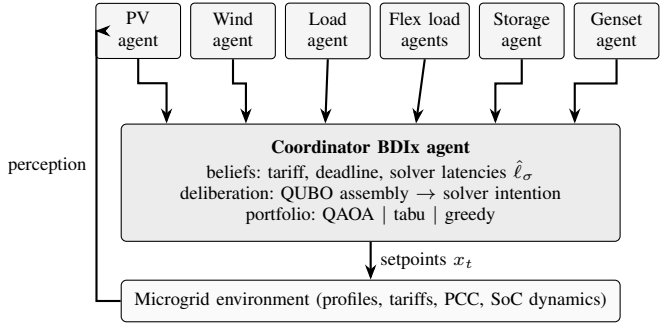

Every agent maintains a belief base, a desire set and an intention store \cite{rao1995,ioannou2020}, with the executable behaviors of the agent organized in a plan library from which intentions are instantiated. Intelligent plan libraries that prioritize desires into intentions from the prevailing beliefs through machine learning and fuzzy logic techniques have been introduced for \bdix{} agents in \cite{ioannou2025planlib}. In the present setting, the solver portfolio of Section~\ref{sec:deliberation} constitutes such a plan library of the coordinator, with the deadline-driven selection rule acting as the desire prioritization mechanism. The reasoning cycle executed at every slot is illustrated in Fig.~\ref{fig:arch}. In the illustrated architecture, beliefs are revised by the field agents through perception and bids are submitted to the coordinator. The per-slot QUBO is then assembled, a solver intention is committed under the prevailing deadline and the committed decision vector is dispatched. The cycle consists of the following phases.
\begin{enumerate}
\item \emph{Perception.} Each field agent revises its beliefs from local measurements. The PV and wind agents observe their instantaneous output, the load agents observe their demand, the storage agent observes its SoC and single-slot charge and discharge feasibility and the coordinator observes the tariff and the prevailing deadline.
\item \emph{Communication.} Each field agent submits a bid that exposes the decision-relevant subset of its beliefs. The storage agent additionally embeds belief-shaped effective prices, as described in Section~\ref{sec:storage}.
\item \emph{Deliberation.} The coordinator assembles the bids into the per-slot QUBO of Section~\ref{sec:qubo}, selects a solver intention under the prevailing deadline as described in Section~\ref{sec:deliberation} and executes it.
\item \emph{Execution.} The committed decision vector is decoded into setpoints that are enacted by the field agents, with the storage agent integrating its SoC dynamics
\begin{equation}
s_{t+1} = s_t + \frac{\Delta t}{E^{\mathrm{s}}}\left( \eta_{\mathrm{c}} P^{\mathrm{s}} x^{\mathrm{c}}_t - \frac{P^{\mathrm{s}}}{\eta_{\mathrm{d}}} x^{\mathrm{d}}_t \right),
\label{eq:soc}
\end{equation}
where $x^{\mathrm{c}}_t$ and $x^{\mathrm{d}}_t$ are the binary charge and discharge decisions.
\item \emph{Belief revision.} The coordinator updates its solver latency beliefs from the measured deliberation time and the storage agent updates its energy valuation, closing the loop.
\end{enumerate}

\section{Methodology}
\label{sec:method}
The methodology of the proposed framework is described in this section. The per-slot dispatch and coalition formation problem is first formulated as a QUBO, the deadline-bounded quantum-classical deliberation mechanism of the coordinator is then described, the belief-shaped storage valuation mechanism is introduced thereafter and the overall execution of the framework is finally summarized in an algorithm.

\subsection{QUBO Formulation of Dispatch and Coalition Formation}
\label{sec:qubo}
The mathematical formulation of the per-slot dispatch and coalition formation problem as a QUBO is described in this subsection. The decision vector, the power balance and cost terms and the settlement of residual imbalances at the PCC are covered.

\subsubsection{Decision Vector}
The dispatch and coalition formation problem of slot $t$ is encoded in the binary vector
\begin{equation}
x = \big[\, x^{\mathrm{g}},\; x^{\mathrm{c}},\; x^{\mathrm{d}},\; u_1,\dots,u_K,\; y_1,\dots,y_{n_I},\; z_1,\dots,z_{n_E} \,\big]^{\!\top}
\end{equation}
of dimension $n = 3 + K + n_I + n_E$, where $x^{\mathrm{g}}$ denotes the genset commitment, $x^{\mathrm{c}}$ and $x^{\mathrm{d}}$ denote the battery charging and discharging decisions, $u_k$ denotes the curtailment of flexible load $k$ and $y_j$ and $z_j$ denote the activation of the import and export blocks. In the evaluated scenario $K=4$ and $n_I = n_E = 2$, hence $n = 11$ and the search space contains $2^{11} = 2048$ configurations per slot. The set of assets activated by $x$ is interpreted as the coalition that serves the slot, with the curtailment flags determining the demand-response sub-coalition.

\subsubsection{Power Balance and Cost}
The net injection implied by $x$ is the affine function $b_0 + a^{\!\top} x$ with $b_0 = R_t - D_t$ and
\begin{align}
a = \big[\, & P^{\mathrm{g}},\, -P^{\mathrm{s}},\, P^{\mathrm{s}},\, P^{\mathrm{f}}_1,\dots,P^{\mathrm{f}}_K, \nonumber\\
& P^{\mathrm{imp}}_1,\dots,P^{\mathrm{imp}}_{n_I},\, -P^{\mathrm{exp}}_1,\dots,-P^{\mathrm{exp}}_{n_E} \,\big]^{\!\top}.
\end{align}
The slot objective combines the linear operating cost $c^{\!\top} x$ with a quadratic balance penalty and exclusion penalties,
\begin{align}
f(x) = \;& \lambda \big( b_0 + a^{\!\top} x \big)^2 + c^{\!\top} x + M\, x^{\mathrm{c}} x^{\mathrm{d}} \nonumber\\
       & + M \sum_{i=1}^{n_I} \sum_{j=1}^{n_E} y_i z_j + m^{\!\top} x ,
\label{eq:obj}
\end{align}
where $\lambda$ is the balance penalty weight in EUR/kW$^2$, $M$ is the mutual exclusion penalty in EUR that prevents simultaneous charging and discharging as well as simultaneous import and export and $m$ is a feasibility mask whose entries take the value $M_{\mathrm{f}} \gg M$ for actions declared infeasible by the storage agent's beliefs. Specifically, charging is declared infeasible when the SoC would exceed $s_{\max}$ within the slot and discharging when the SoC would fall below $s_{\min}$. In the considered model the mask is populated only by the storage agent. The linear cost vector $c$ collects the per-slot fuel cost $c^{\mathrm{g}} P^{\mathrm{g}} \Delta t$, the effective storage prices $c_{\mathrm{C},t} P^{\mathrm{s}} \Delta t$ and $c_{\mathrm{D},t} P^{\mathrm{s}} \Delta t$ of Section~\ref{sec:storage}, the discomfort costs $c^{\mathrm{f}}_k P^{\mathrm{f}}_k \Delta t$, the import costs $\pi_t P^{\mathrm{imp}}_j \Delta t$ and the export revenues $-\pi^{\mathrm{exp}} P^{\mathrm{exp}}_j \Delta t$. It is noted that subscripted or superscripted $c$ terms denote scalar cost coefficients, whereas $c$ without a subscript in $c^\top x$ denotes the linear cost vector.

Using $x_i^2 = x_i$, the objective in \eqref{eq:obj} is rewritten exactly in QUBO form,
\begin{equation}
\min_{x \in \{0,1\}^n} \; x^{\!\top} Q\, x + c_0,
\qquad c_0 = \lambda b_0^2,
\label{eq:qubo}
\end{equation}
with $Q = \lambda\, a a^{\!\top} + \Delta$, where the diagonal correction $\Delta_{ii} = 2 \lambda b_0 a_i + c_i + m_i$ absorbs the linear terms and each exclusion penalty $M$ is split as $M/2$ into the two corresponding symmetric off-diagonal entries, so that the quadratic form $x^{\!\top} Q\, x$ reproduces the penalty terms of \eqref{eq:obj} exactly. The formulation \eqref{eq:qubo} maps directly to the Ising Hamiltonians accepted by quantum annealers and by QAOA \cite{lucas2014,glover2019}.

\subsubsection{Residual Settlement}
Let $x^\star$ denote the committed solution and $\rho = b_0 + a^{\!\top} x^\star$ the residual imbalance. A residual surplus $\rho > 0$ is spilled renewable energy. A residual deficit $\rho<0$, which arises because grid exchange is encoded in discrete blocks, is settled physically by the continuous PCC regulation slack up to $P^{\mathrm{slk}}$ at the premium tariff $\kappa \pi_t$ and only the remainder $\max(0, -\rho - P^{\mathrm{slk}})$ is accounted as unserved energy at $c_{\mathrm{VoLL}}$. This settlement reflects the behavior of a physical PCC and decouples the modeling granularity of the binary encoding from the reliability metrics.

\subsection{Deadline-Bounded Quantum-Classical Deliberation}
\label{sec:deliberation}
The deadline-bounded deliberation mechanism of the coordinator agent is described in this subsection, comprising the portfolio of quantum, quantum-inspired and classical solvers and the belief-driven selection of the solver intention at every slot.

\subsubsection{Solver Portfolio}
The coordinator maintains a portfolio of solvers sharing the interface $(x^\star, E^\star) = \sigma(Q, c_0)$, where $E^\star$ denotes the objective value of the returned bitstring $x^\star$, which realizes the plan library of its \bdix{} architecture \cite{ioannou2025planlib}. The deliberation portfolio admitted by the selection rule in \eqref{eq:select} comprises QAOA, tabu search and greedy descent, whereas simulated annealing, binary PSO and exhaustive enumeration are maintained only as benchmarking baselines that are executed on the identical QUBO instances for evaluation purposes.

\paragraph{QAOA}
The quantum deliberation action prepares the variational state \cite{farhi2014}
\begin{equation}
\ket{\psi(\boldsymbol{\gamma},\boldsymbol{\beta})} = \prod_{l=1}^{p} e^{-i \beta_l H_M}\, e^{-i \gamma_l H_C} \ket{+}^{\otimes n},
\label{eq:qaoa}
\end{equation}
where $H_C$ is the diagonal cost Hamiltonian whose spectrum is the min-max normalized QUBO energy landscape of \eqref{eq:qubo}, $H_M = \sum_{i=1}^{n} X_i$ is the transverse-field mixer and $p$ is the circuit depth. The angles $(\boldsymbol{\gamma},\boldsymbol{\beta})$ are optimized by Nelder-Mead search on the energy expectation $\langle \psi | H_C | \psi \rangle$ and the dispatch solution is the lowest-energy bitstring among the $M_{\mathrm{top}}$ most probable measurement outcomes, of which the $K_{\mathrm{pol}}$ best candidates are refined by a steepest-descent polishing stage. The min-max normalization rescales the QUBO energies to the unit interval and does not change the minimizing bitstring. In this paper the state \eqref{eq:qaoa} is evolved by exact statevector simulation, which is tractable at the evaluated problem size and renders the latency and quality statistics fully reproducible; the identical interface accommodates gate-based hardware backends or annealing devices without modification of the agent layer.

\paragraph{Classical and quantum-inspired baselines}
Multi-restart single-flip tabu search \cite{glover1990}, which mirrors the quantum-inspired QUBO solver shipped with commercial quantum toolchains, is also included in the deliberation portfolio together with steepest-descent greedy search as a sub-millisecond fallback. The benchmarking baselines comprise simulated annealing \cite{kirkpatrick1983} emulating an annealing schedule, binary PSO \cite{kennedy1997} connecting the framework to the swarm-based variant of \bdix{} orchestration \cite{ioannou2022access} and exhaustive enumeration, which is tractable for $n \le 16$ and provides the exact optimum used as ground truth in the evaluation. Single-flip moves are evaluated in $O(n)$ through the standard delta relation $\delta_i = (1-2x_i)\big( Q_{ii} + 2 \sum_{j \ne i} Q_{ij} x_j \big)$.

\subsubsection{Latency Beliefs and Intention Selection}
For every solver $\sigma$ the coordinator maintains the latency belief $\hat{\ell}_\sigma$, initialized by an optimistic prior and revised after every invocation by the exponential moving average
\begin{equation}
\hat{\ell}_\sigma \leftarrow (1-\alpha_\ell)\, \hat{\ell}_\sigma + \alpha_\ell\, \ell_\sigma,
\label{eq:ema}
\end{equation}
where $\ell_\sigma$ is the measured deliberation time and $\alpha_\ell = 0.4$. Given the deadline $\Delta^{\mathrm{dl}}_t$ perceived for slot $t$, the solver intention is committed by the guarded preference rule
\begin{equation}
\sigma_t =
\begin{cases}
\text{QAOA}, & n \le n_{\max} \;\wedge\; \hat{\ell}_{\mathrm{qaoa}} \le \theta_1 \Delta^{\mathrm{dl}}_t,\\
\text{tabu}, & \hat{\ell}_{\mathrm{tabu}} \le \theta_2 \Delta^{\mathrm{dl}}_t,\\
\text{greedy}, & \text{otherwise},
\end{cases}
\label{eq:select}
\end{equation}
with safety margins $\theta_1 = 0.8$ and $\theta_2 = 0.9$ and with $n_{\max}$ bounding the problem size admitted to the simulated quantum backend. The rule \eqref{eq:select} realizes graceful degradation: the highest-quality admissible solver is preferred and the portfolio is descended whenever the learned latency beliefs indicate that the deadline would be jeopardized.

\subsection{Belief-Shaped Storage Valuation}
\label{sec:storage}
The belief-shaped storage valuation mechanism through which the future value of stored energy is injected into the otherwise myopic per-slot optimization is described in this subsection. The per-slot QUBO \eqref{eq:qubo} is myopic and, if the storage actions were priced at the physical degradation cost alone, charging would never be selected, since the immediate cost of absorbing $P^{\mathrm{s}}$ is positive whereas the benefit materializes only in later slots. Rather than expanding the optimization horizon, which would multiply the number of qubits by the lookahead length, the intertemporal information is injected by the storage agent itself through belief revision.

At every slot the storage agent values its stored energy at the discounted future-peak price
\begin{equation}
v_t = \delta\, \eta_{\mathrm{RT}} \max_{\tau \in (t,\, t+H]} \pi_\tau,
\label{eq:val}
\end{equation}
where $\eta_{\mathrm{RT}} = \eta_{\mathrm{c}} \eta_{\mathrm{d}}$ is the round-trip efficiency, $H$ is the lookahead window and $\delta \in (0,1]$ is a discount reflecting forecast uncertainty and the finite probability that the peak can be captured. For slots near the end of the horizon the window is truncated to $(t,\, \min(t+H,\, T)]$. The effective charge and discharge prices embedded into the bid are
\begin{align}
c_{\mathrm{C},t} &= \mathrm{clip}\big( c_{\mathrm{deg}} - v_t + \mu (s_t - s_{\mathrm{ref}}) \big), \nonumber\\
c_{\mathrm{D},t} &= \mathrm{clip}\big( c_{\mathrm{deg}} + v_t - \mu (s_t - s_{\mathrm{ref}}) \big),
\label{eq:cd}
\end{align}
where $\mu$ weights a mild SoC-reference shaping term, $s_{\mathrm{ref}}$ is the reference SoC and $\mathrm{clip}(q) = \min(\max(q, q_{\min}), q_{\max})$ bounds the prices to the safe interval $[q_{\min}, q_{\max}]$. The values of $q_{\min}$ and $q_{\max}$ are listed in Table~\ref{tab:params}. A negative $c_{\mathrm{C},t}$ expresses that charging is presently profitable net of its future value. Charging is thereby made attractive in the myopic QUBO when cheap energy is available, namely during off-peak tariffs or renewable surplus, whereas the elevated $c_{\mathrm{D},t}$ outside peak periods expresses the opportunity cost of releasing energy early. The realized cost accounting retains the physical degradation price, hence the shaping terms influence only the decision and not the reported economics.

\subsection{Overall Execution of the Framework}
\label{sec:overall}
The overall execution of the proposed framework is summarized in Algorithm~\ref{alg:cycle}, in which the mechanisms of Sections~\ref{sec:qubo} to~\ref{sec:storage} are composed into the per-slot reasoning cycle of the \bdix{} agents. At every slot the beliefs of the field agents are revised from local measurements in the perception phase (line 2) and the effective charge and discharge prices of the storage agent are computed from the belief-shaped valuation of \eqref{eq:val} and \eqref{eq:cd} (line 3). The decision-relevant beliefs are then exposed to the coordinator through bids in the communication phase (line 4) and the per-slot QUBO of \eqref{eq:qubo} is assembled from these bids (line 5). The deliberation phase follows: the prevailing deadline is perceived and the solver intention is committed by the guarded preference rule \eqref{eq:select} (line 6), after which the committed solver is executed on the assembled QUBO and its deliberation latency is measured (line 7). The measured latency is folded into the corresponding latency belief by the exponential moving average \eqref{eq:ema} (line 8), through which the deadline awareness of subsequent slots is sustained. In the execution phase the residual imbalance is settled at the PCC, the decoded setpoints are enacted by the field agents (line 9) and the SoC dynamics of \eqref{eq:soc} are integrated by the storage agent (line 10), closing the loop for the next slot. The cycle is repeated for all $T$ slots of the horizon and no step requires global state beyond the exchanged bids, hence the distributed character of the DAI framework is preserved.

\begin{algorithm}[!t]
\caption{Per-Slot Quantum-Assisted \bdix{} EMS Cycle}
\label{alg:cycle}
\begin{algorithmic}[1]
\FOR{$t = 1$ \TO $T$}
\STATE every field agent revises its beliefs (perception)
\STATE storage agent computes $c_{\mathrm{C},t}, c_{\mathrm{D},t}$ by \eqref{eq:val}--\eqref{eq:cd}
\STATE bids are submitted to the coordinator (communication)
\STATE coordinator assembles $Q, c_0$ by \eqref{eq:qubo}
\STATE coordinator perceives $\Delta^{\mathrm{dl}}_t$ and selects $\sigma_t$ by \eqref{eq:select}
\STATE $(x^\star, E^\star) \leftarrow \sigma_t(Q, c_0)$; latency $\ell_{\sigma_t}$ is measured
\STATE latency belief $\hat{\ell}_{\sigma_t}$ is revised by \eqref{eq:ema}
\STATE residual $\rho$ is settled at the PCC; setpoints are enacted
\STATE storage agent integrates \eqref{eq:soc} (execution)
\ENDFOR
\end{algorithmic}
\end{algorithm}
\section{Performance Evaluation}
\label{sec:eval}
The performance evaluation of the proposed framework is presented in this section, covering the simulation setup, the dispatch behavior, the solver benchmark together with the deadline adherence, a comparison with state-of-the-art approaches, an ablation study and a discussion of scope and limitations.

\subsection{Setup}
The framework was implemented as a self-contained MATLAB simulation comprising the agent classes, the QUBO assembly and the six solvers, requiring no toolboxes; the QAOA backend is an exact statevector simulator with the mixer applied by per-qubit tensor contraction. The scenario parameters are listed in Table~\ref{tab:params}. All random streams are seeded. Therefore, every reported figure is exactly reproducible under the evaluated MATLAB release; the seeded realization is implementation dependent and differs under other interpreters of the same code. The critical flag is drawn with probability $p_{\mathrm{c}} = 0.20$ per slot. The final seeded realization used in the reported results produces 27 critical slots and 69 routine slots. In the per-slot benchmark, all six solvers are executed on the identical QUBO of every slot, while the dispatch follows the solver committed by the deliberation rule \eqref{eq:select}. The dispatch quality of every approach is measured on the committed decision vectors against the exhaustive optimum, whereas the benchmark characterizes independent invocations of each solver. The baseline family and ablation runs are executed by the same simulator under the exogenous profiles of the final trace, with the stochastic solver streams reseeded by the executing platform.

\begin{table}[!t]
\caption{Simulation Parameters}
\label{tab:params}
\centering
\footnotesize
\setlength{\tabcolsep}{2.5pt}
\begin{tabular}{l l}
\toprule
Parameter & Value \\
\midrule
Horizon, slot length & $T=96$, $\Delta t = 15$ min \\
PV, wind capacity & 40 kWp, 15 kW \\
Inflexible peak demand & 40 kW \\
Flexible loads $P^{\mathrm{f}}_k$ & 3, 4, 5 and 6 kW \\
Discomfort prices $c^{\mathrm{f}}_k$ & 0.40, 0.50, 0.60, 0.80 EUR/kWh \\
BESS & 20 kW / 60 kWh, $\eta_{\mathrm{c}}=\eta_{\mathrm{d}}=0.95$ \\
SoC limits, $c_{\mathrm{deg}}$ & $[0.10,\,0.95]$, 0.05 EUR/kWh \\
Genset & 30 kW at 0.35 EUR/kWh \\
Import and export blocks & $\{10, 20\}$ kW each \\
Tariff $\pi_t$ (off, day, peak) & 0.15, 0.22, 0.30 EUR/kWh \\
Feed-in tariff, slack, $\kappa$ & 0.08 EUR/kWh, 10 kW, 1.25 \\
$c_{\mathrm{VoLL}}$, $\lambda$, $M$ & 1.50 EUR/kWh, 0.08 EUR/kW$^2$, 50 EUR \\
Deadlines $\Delta^{\mathrm{dl}}_{\mathrm{n}}, \Delta^{\mathrm{dl}}_{\mathrm{c}}$ & 300 ms, 60 ms \\
Critical-slot realization & 27 critical and 69 routine slots in the final trace \\
QAOA ansatz & $p=2$, 100 evals, $n_{\max}=12$ \\
QAOA sampling, polish & $M_{\mathrm{top}}=1024$, $K_{\mathrm{pol}}=32$ \\
Storage valuation & $H=48$, $\delta=0.85$, $\mu=0.10$, $s_{\mathrm{ref}}=0.5$ \\
Price clip $q_{\min}$, $q_{\max}$ & $-0.30$, $0.50$ EUR/kWh \\
\bottomrule
\end{tabular}
\end{table}

\subsection{Dispatch Behavior}
\begin{figure}[!t]
\centering
\includegraphics[width=\columnwidth]{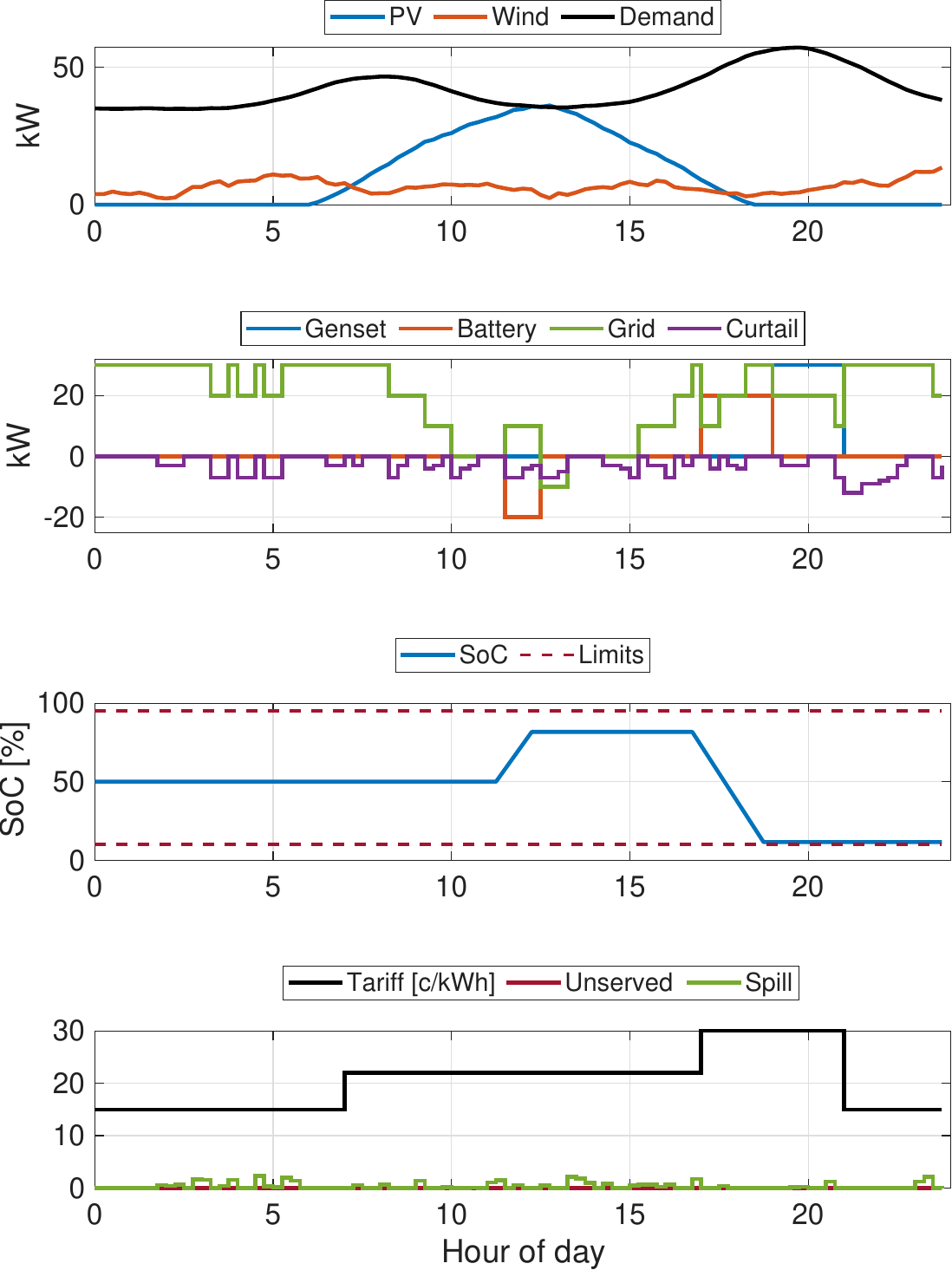}
\caption{24-hour operation of the proposed framework: sources and consumers, dispatch decisions, battery SoC and tariff together with the unserved and spilled power traces.}
\label{fig:dispatch}
\end{figure}

Fig.~\ref{fig:dispatch} presents the 24-hour operation, showing from top to bottom the fluctuating sources and consumers, the committed dispatch decisions, the battery SoC and the tariff together with unserved and spilled power. All power traces are expressed in kW, the SoC in percent and the tariff in EUR cents per kWh. The committed coalitions track the fluctuating sources and consumers closely: grid import blocks and short genset commitments cover the morning ramp, the midday PV surplus is absorbed partly by charging and demand response is activated predominantly when the tariff or the imbalance renders curtailment cheaper than additional supply. The storage trajectory demonstrates the effect of the belief-shaped valuation of Section~\ref{sec:storage}: the battery remains close to 50 percent SoC overnight, charges during the midday PV-surplus interval from 11:30 to 12:15 up to 81.7 percent SoC and discharges from 17:00 to 18:45 down to 11.5 percent SoC. It is noted that instantaneous power traces in kW are shown in Fig.~\ref{fig:dispatch}, whereas integrated energy totals in kWh are reported in this paragraph. No unserved energy is recorded after demand response and PCC slack settlement. The final trace includes 65.25 kWh of curtailed flexible demand, 15.55 kWh of PCC regulation slack and 8.75 kWh of spilled renewable energy. Renewable energy utilization reaches 97.83 percent and the daily operating cost is 146.24 EUR, with the genset committed for 2.0 h.

\subsection{Solver Benchmark and Deadline Adherence}
\begin{figure*}[!t]
\centering
\includegraphics[width=0.92\textwidth]{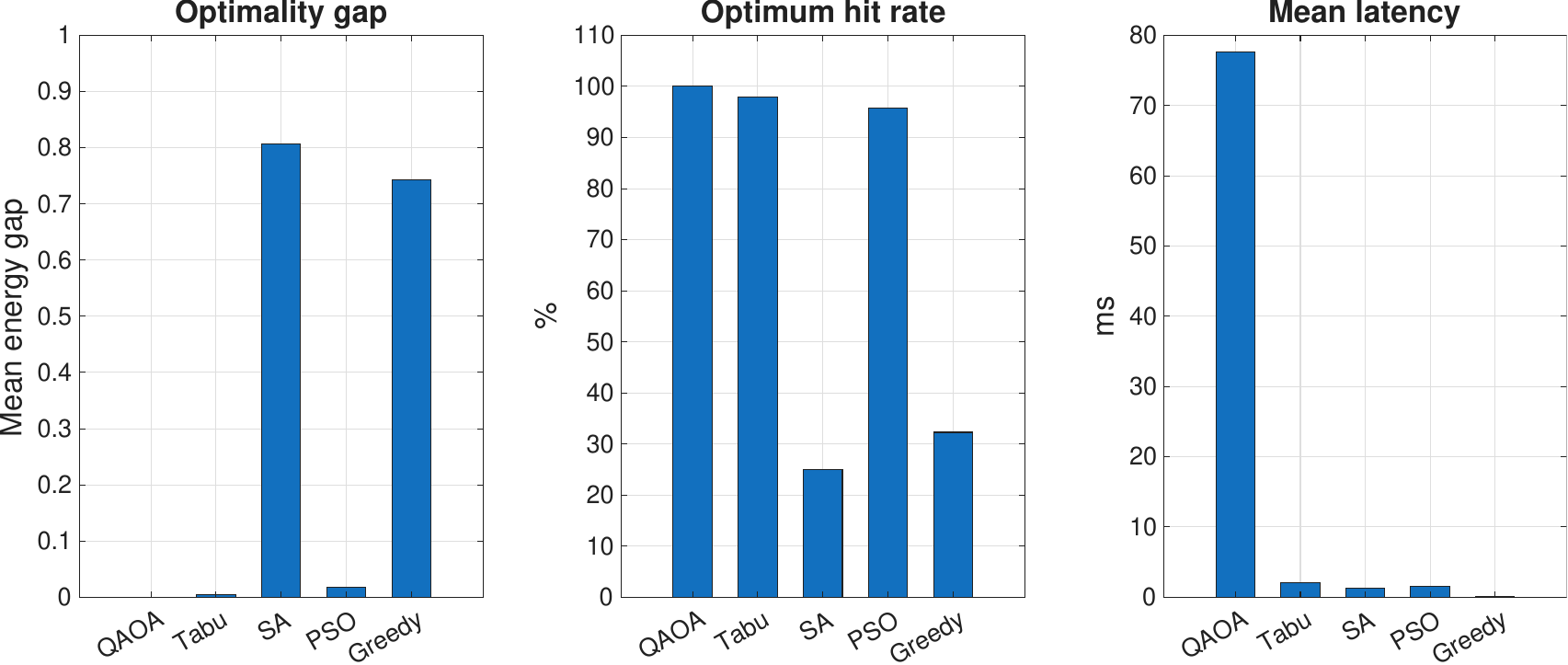}
\caption{Solver benchmark over the 96 per-slot QUBO instances, where QAOA, Tabu, SA, PSO and Greedy denote hybrid QAOA, tabu search, simulated annealing, binary PSO and greedy descent.}
\label{fig:solvers}
\end{figure*}

Table~\ref{tab:solvers} and Fig.~\ref{fig:solvers} report the per-slot benchmark of all solvers against exhaustive enumeration over the 96 identical QUBO instances. The mean optimality gap, the exact-optimum hit rate and the mean latency are depicted in Fig.~\ref{fig:solvers}. The hybrid QAOA action remains the committed routine-slot solver and its mean committed latency in the final trace is 79.8 ms. The committed dispatch of the deadline-bounded portfolio uses QAOA on 69 routine slots and tabu search on 27 critical slots; the hit rate of tabu search reported in Table~\ref{tab:solvers} refers to independent benchmark invocations whose random restarts differ from those of the committed instances, whereas the committed tabu instances of the deadline-bounded trace reached the exhaustive optimum in every critical slot. The classical polish is essential for the solution quality of the hybrid action: pure sampling of the variational state without refinement attains a markedly lower hit rate, because the shallow $p=2$ ansatz under a bounded angle optimization budget concentrates probability mass on the basins of the low-energy states rather than on the exact ground state bitstring and the steepest-descent polish recovers the minimizer of every sampled basin at a cost in the microsecond range. Tabu search and binary PSO attain high quality at lower latency, which is the expected outcome at $n=11$ where the simulated statevector evolution dominates the QAOA runtime; the value of the quantum action lies in its scaling prospects on hardware backends, which the agentic interface admits without modification. Simulated annealing under the evaluated budget and greedy descent are dominated. The weaker solution quality of greedy descent is offset by its sub-millisecond latency, through which deadline feasibility is maintained in any evaluated regime. Regarding the time consumed by deliberation, the mean QAOA latency occupies 26.6 percent of the routine deadline and 33.2 percent of the guarded budget $\theta_1 \Delta^{\mathrm{dl}}_{\mathrm{n}}$, the maximum observed latency is 225.9 ms and the mean tabu latency occupies 3.3 percent of the critical deadline. The cumulative deliberation time across the entire day is 5.56 s, which corresponds to 0.0064 percent of the 24-hour horizon.

\begin{table}[!t]
\caption{Per-Slot Solver Benchmark (96 Instances, $n=11$)}
\label{tab:solvers}
\centering
\footnotesize
\setlength{\tabcolsep}{4pt}
\begin{tabular}{l c c c}
\toprule
Solver & Mean gap & Opt. hit rate & Latency \\
\midrule
Hybrid QAOA (sim.) \cite{farhi2014,koretsky2021} & 0.000 & 100.0\,\% & 77.67 ms \\
Tabu search \cite{glover1990,glover2019} & 0.005 & 97.9\,\% & 2.07 ms \\
Simulated annealing \cite{kirkpatrick1983,morstyn2023} & 0.807 & 25.0\,\% & 1.26 ms \\
Binary PSO \cite{kennedy1997,radosavljevic2016} & 0.018 & 95.8\,\% & 1.50 ms \\
Greedy descent & 0.743 & 32.3\,\% & 0.06 ms \\
Exhaustive (exact) & 0.000 & 100.0\,\% & 0.33 ms \\
\bottomrule
\end{tabular}
\end{table}

\begin{table*}[!t]
\caption{Comparison of the Proposed Framework With State-of-the-Art Approaches (Best Values in Bold; the DRL Family Is Included Qualitatively)}
\label{tab:sota}
\centering
\footnotesize
\setlength{\tabcolsep}{2.3pt}
\begin{tabular}{l l c c c c c c c c}
\toprule
Approach & Core method & Agentic & Quantum & Deadline aware & Cost [EUR]$^{\S}$ & Misses & Mean gap & Hit [\%] & Energy [J] \\
\midrule
PSO EMS \cite{radosavljevic2016} & PSO & No & No & No & 147.14 & \textbf{0} & 0.018 & 95.80 & 2.16 \\
MAS real-time EMS \cite{logenthiran2012} & Rule-based heuristics & Yes & No & No & 203.36 & \textbf{0} & 0.743 & 32.30 & \textbf{0.09} \\
DRL EMS \cite{du2020} & Deep RL & No & No & Implicit$^{\dagger}$ & \multicolumn{5}{c}{not evaluated$^{\ddagger}$} \\
Quantum distributed UC \cite{nikmehr2022} & Quantum annealing (SA emul.) & No & Yes & No & 222.39 & \textbf{0} & 0.807 & 25.00 & 1.81 \\
QAOA UC \cite{koretsky2021} & Hybrid QAOA, unconditional & No & Yes & No & 147.42 & 27 & \textbf{0.000} & \textbf{100.00} & 111.85 \\
Annealing OPF \cite{morstyn2023} & Quantum annealing (SA emul.) & No & Yes & No & 222.39 & \textbf{0} & 0.807 & 25.00 & 1.81 \\
DAI nano-grid \cite{ioannou2024access} & \bdix{} plans with PSO & Yes & No & No & 147.14 & \textbf{0} & 0.018 & 95.80 & 2.16 \\
Proposed & \bdix{} portfolio, deadline bounded & Yes & Yes & Yes & \textbf{146.24} & \textbf{0} & \textbf{0.000} & \textbf{100.00} & 83.38 \\
\bottomrule
\multicolumn{10}{@{}p{0.97\textwidth}@{}}{\footnotesize $^{\dagger}$Trained DRL policies deliver constant-time inference, hence deadlines are met implicitly without deadline-aware deliberation. $^{\ddagger}$The DRL family was not re-implemented because it requires training under a protocol that differs from the evaluated one and is therefore reported qualitatively. $^{\S}$Family costs are measured by executing the same simulator under the exogenous profiles of the final trace; the greedy based family is deterministic, whereas the costs of the stochastic families reflect the seeded solver streams of the executing platform. The mean gap and hit rate columns reproduce the independent per-slot benchmark of Table~\ref{tab:solvers}, whereas the cost, miss and energy columns are measured from the family executions, which explains why a family may attain a 100 percent benchmark hit rate while its family cost differs from the lower bound. Deadline misses are counted from the latencies of the evaluated MATLAB platform and deadline violations do not alter the realized operating cost, which the simulator accounts independently of timing. Zero misses alone do not imply deadline-aware solver selection, since only the proposed framework reasons about deliberation latency.}
\end{tabular}
\end{table*}

\begin{figure}[!t]
\centering
\includegraphics[width=\columnwidth]{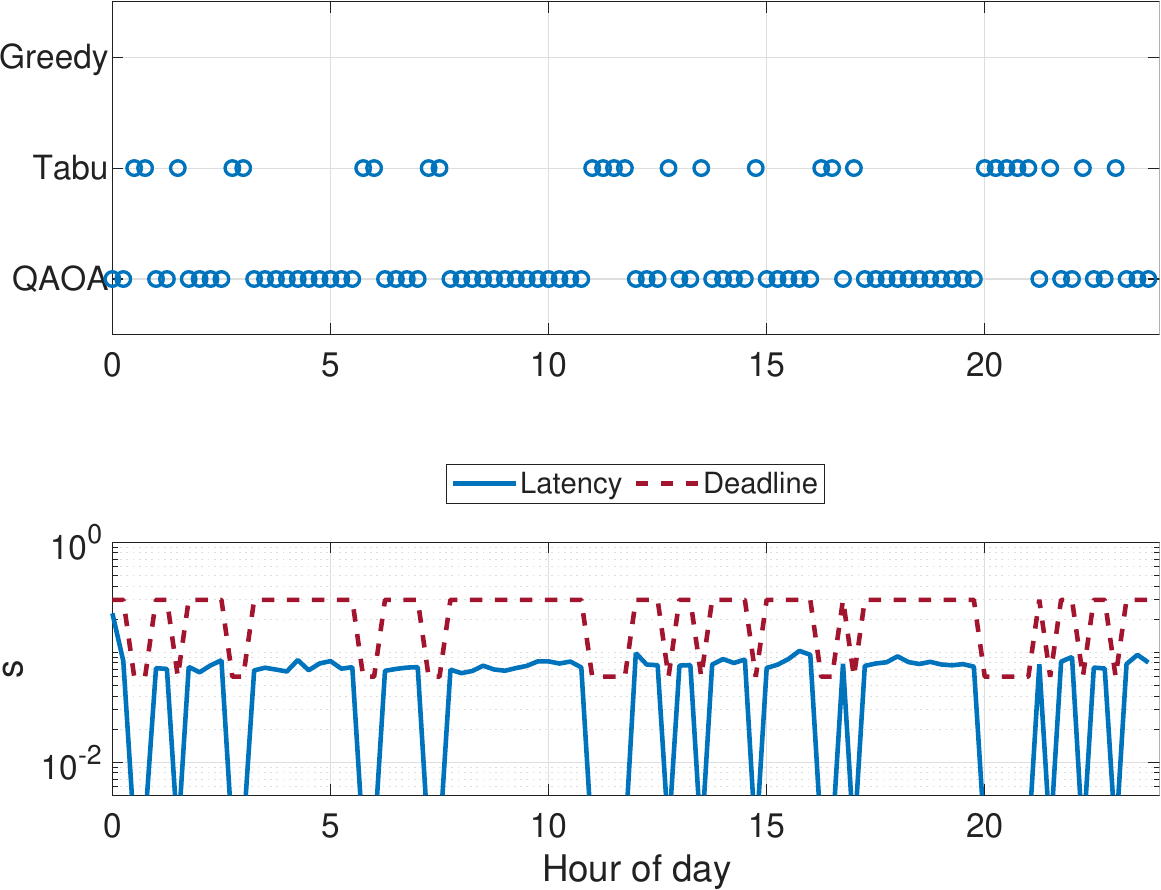}
\caption{Deadline-bounded meta-deliberation over the 24-hour horizon, with the latency shown in seconds on a logarithmic scale.}
\label{fig:deadline}
\end{figure}

The deadline adherence of the deliberation mechanism is examined next. The solver intention committed in every slot, degrading from QAOA to tabu search on the critical slots, is depicted in the top panel of Fig.~\ref{fig:deadline} and the measured deliberation latency against the per-slot deadline is depicted in the bottom panel. The coordinator commits QAOA on all 69 routine slots and degrades to tabu search on exactly the 27 critical slots whose 60 ms deadline excludes the simulated quantum action, with the measured latency dropping accordingly to approximately 2 ms. Zero deadlines are missed across the entire horizon and the greedy fallback is never required. The latency beliefs converge within the first slots of each regime, confirming that the exponential moving average \eqref{eq:ema} suffices for the stationary latency processes of the evaluated backends.

\begin{table}[!t]
\caption{Ablation of the Proposed Mechanisms}
\label{tab:ablation}
\centering
\footnotesize
\setlength{\tabcolsep}{2.5pt}
\begin{tabular}{l c c c}
\toprule
Configuration & Cost [EUR] & Unserved [kWh] & Genset [h] \\
\midrule
Full framework & 146.24 & 0 & 2.0 \\
No storage valuation & 152.75 & 0 & 3.5 \\
No valuation, no PCC slack & 172.77 & 16.15 & 3.5 \\
\bottomrule
\end{tabular}
\end{table}

\subsection{Comparison With the State of the Art, Ablation and Discussion}
Table~\ref{tab:sota} compares the proposed framework with representative state-of-the-art approaches that address dispatch and commitment under the evaluated scenario. A direct comparison of the costs reported by these works is not meaningful because the underlying scenarios, asset fleets and tariffs differ; therefore the algorithmic core of each family was re-implemented and the complete 24-hour scenario was dispatched again with each method unconditionally in every slot, so that all quantitative columns of Table~\ref{tab:sota} are measured on identical instances. Binary PSO represents the metaheuristic EMS family \cite{radosavljevic2016} and the plan library of the DAI nano-grid framework \cite{ioannou2024access}, simulated annealing represents the schedule of annealing-based quantum devices \cite{morstyn2023,nikmehr2022}, QAOA represents the gate-based variational family \cite{koretsky2021} and greedy descent represents the rule-based reactive MAS family \cite{logenthiran2012}; the DRL family \cite{du2020} was not re-implemented and is reported qualitatively. The measurements expose the central trade-off. The exact per-slot lower bound of 146.24 EUR is established by the exhaustive enumeration of Table~\ref{tab:solvers}, which remains tractable only at the evaluated problem size. Exhaustive enumeration is used only as a ground truth generator for this small problem size and not as a scalable EMS method. Among the unconditional families, only the hybrid QAOA family violates the critical deadlines, on all 27 critical slots of the final trace, since the committed statevector latency never falls below 64.6 ms, whereas the annealing-emulated and classical families operate well inside every deadline at the evaluated problem size. Unconditional hybrid QAOA reaches 147.42 EUR while violating all 27 critical deadlines, the annealing schedule degrades to 222.39 EUR under the evaluated iteration budget and the PSO based families operate at 147.14 EUR without deadline awareness or readiness for quantum hardware. The rule-based family meets every deadline but at 203.36 EUR, which is 39 percent above the lower bound. The proposed deadline-bounded deliberation reduces this tradeoff by combining low cost with zero deadline misses: a cost of 146.24 EUR is attained, matching the exact lower bound, with zero deadline misses, a mean dispatch gap of 0.000 and a 100.0 percent optimum hit rate on the committed decision vectors. The readiness for quantum hardware, which the classical families lack, is retained. It is also the only approach in which the solver selection is performed by the agents themselves. Regarding the computational energy required for deliberation, the Energy column of Table~\ref{tab:sota} is estimated as the measured active single-core compute time multiplied by a nominal active core power of 15 W, with the cryogenic and control overheads of physical quantum hardware not modeled. The proposed framework consumes 83.38 J in the final trace, the unconditional hybrid QAOA family consumes 111.85 J and the remaining families consume between 0.09 J and 2.16 J, hence the deadline-bounded portfolio also reduces the deliberation energy relative to unconditional quantum execution. All of these figures are negligible relative to the dispatched energy, since the daily served demand of approximately 1 MWh corresponds to 3.6 GJ, which is seven orders of magnitude above the deliberation energy. Hence, the computational energy does not influence the choice among the families on simulated backends. It would, however, become a meaningful belief of the coordinator once physical quantum hardware and its cryogenic overheads are included and its inclusion in the deliberation rule is noted as future work.

Table~\ref{tab:ablation} quantifies the contribution of the proposed mechanisms by deactivating them cumulatively under the identical scenario seed. Without the belief-shaped storage valuation the battery is operated on the degradation price alone, never charges and the daily cost rises from 146.24 to 152.75 EUR, an increase of 4.5 percent, with the genset run time extended to 3.5 h. Without the PCC slack settlement, residual block-granularity deficits are additionally booked as unserved energy at $c_{\mathrm{VoLL}}$ and the apparent cost rises to 172.77 EUR with 16.15 kWh of unserved energy, illustrating that the settlement layer is required for the binary block encoding to coexist with reliability accounting. It is recalled that all reported costs are computed with the physical degradation price, hence the belief-shaped prices of Section~\ref{sec:storage} influence only the decisions and not the cost accounting.

Several observations on the scope and limitations of the evaluation are warranted. The evaluated problem size of $n=11$ binaries per slot is deliberately small so that exhaustive ground truth is available for every instance, which renders the quality statistics exact rather than estimated. At this size the simulated QAOA cannot outperform tuned classical heuristics in latency and no such claim is made; the framework is positioned as ready for quantum hardware deployment and the contribution resides in the agentic mechanism through which it is decided when a quantum action is admissible under the prevailing deadline. Larger microgrids increase $n$ linearly with the number of controllable assets and the per-feeder decomposition that is natural in the agentic setting bounds the qubit requirement of each subproblem, which, together with hardware backends, constitutes the intended scaling path. The storage valuation \eqref{eq:val} relies on a tariff forecast that is exact in the evaluated setting; forecast errors would propagate into the effective prices. The robust treatment of those errors, for example through the discount $\delta$ or through distributional beliefs, is left for future work.

\section{Conclusion}
\label{sec:conclusion}
The concluding remarks of the paper and the directions for future research are presented in this section. A quantum-assisted DAI framework, realized by \bdix{} agents, has been proposed for the real-time orchestration of hybrid renewable microgrids. In the proposed framework, per-slot dispatch and coalition formation are assembled as QUBO problems from distributed agent bids and quantum, quantum-inspired and classical solvers constitute alternative deliberation actions that are selected per slot under hard deadlines through learned latency beliefs. A belief-shaped storage valuation mechanism has been introduced that injects the future value of energy into the myopic per-slot optimization through the storage agent's effective prices. In a fully reproducible 24-hour evaluation the framework recorded zero unserved energy after demand response and PCC slack settlement, reached 97.83 percent renewable utilization, operated at a daily cost of 146.24 EUR and missed zero deliberation deadlines while degrading gracefully from the hybrid QAOA action on 69 routine slots to tabu search on 27 critical slots. The ablation showed that deactivating the storage valuation increases the daily operating cost from 146.24 EUR to 152.75 EUR, corresponding to a 4.5 percent increase. Future work includes the execution of the QAOA action on gate-based hardware and annealing backends through the preserved solver interface, per-feeder decomposition for larger asset counts, coupling of the supervisory layer to electromagnetic-transient simulation and integration of large language model assisted intent translation into the agents' desire formation.

\bibliographystyle{IEEEtran}
\bibliography{references}

@INPROCEEDINGS{lasseter2002,
  author    = {R. H. Lasseter},
  title     = {{MicroGrids}},
  booktitle = {Proc. IEEE Power Eng. Soc. Winter Meeting},
  address   = {New York, NY, USA},
  volume    = {1},
  pages     = {305--308},
  year      = {2002},
  doi       = {10.1109/PESW.2002.985003},
  note    = {doi: 10.1109/PESW.2002.985003}
}

@ARTICLE{olivares2014,
  author  = {D. E. Olivares and A. Mehrizi-Sani and A. H. Etemadi and C. A. Ca{\~n}izares and R. Iravani and M. Kazerani and A. H. Hajimiragha and O. Gomis-Bellmunt and M. Saeedifard and R. Palma-Behnke and G. A. Jim{\'e}nez-Est{\'e}vez and N. D. Hatziargyriou},
  title   = {Trends in microgrid control},
  journal = {IEEE Trans. Smart Grid},
  volume  = {5},
  number  = {4},
  pages   = {1905--1919},
  month   = jul,
  year    = {2014},
  doi     = {10.1109/TSG.2013.2295514},
  note    = {doi: 10.1109/TSG.2013.2295514}
}

@ARTICLE{zia2018,
  author  = {M. F. Zia and E. Elbouchikhi and M. Benbouzid},
  title   = {Microgrids energy management systems: {A} critical review on methods, solutions, and prospects},
  journal = {Appl. Energy},
  volume  = {222},
  pages   = {1033--1055},
  month   = jul,
  year    = {2018},
  doi     = {10.1016/j.apenergy.2018.04.103},
  note    = {doi: 10.1016/j.apenergy.2018.04.103}
}

@ARTICLE{dimeas2005,
  author  = {A. L. Dimeas and N. D. Hatziargyriou},
  title   = {Operation of a multiagent system for microgrid control},
  journal = {IEEE Trans. Power Syst.},
  volume  = {20},
  number  = {3},
  pages   = {1447--1455},
  month   = aug,
  year    = {2005},
  doi     = {10.1109/TPWRS.2005.852060},
  note    = {doi: 10.1109/TPWRS.2005.852060}
}

@ARTICLE{logenthiran2012,
  author  = {T. Logenthiran and D. Srinivasan and A. M. Khambadkone and H. N. Aung},
  title   = {Multiagent system for real-time operation of a microgrid in real-time digital simulator},
  journal = {IEEE Trans. Smart Grid},
  volume  = {3},
  number  = {2},
  pages   = {925--933},
  month   = jun,
  year    = {2012},
  doi     = {10.1109/TSG.2012.2189028},
  note    = {doi: 10.1109/TSG.2012.2189028}
}

@ARTICLE{mcarthur2007,
  author  = {S. D. J. McArthur and E. M. Davidson and V. M. Catterson and A. L. Dimeas and N. D. Hatziargyriou and F. Ponci and T. Funabashi},
  title   = {Multi-agent systems for power engineering applications, {Part I}: {Concepts}, approaches, and technical challenges},
  journal = {IEEE Trans. Power Syst.},
  volume  = {22},
  number  = {4},
  pages   = {1743--1752},
  month   = nov,
  year    = {2007},
  doi     = {10.1109/TPWRS.2007.908471},
  note    = {doi: 10.1109/TPWRS.2007.908471}
}

@ARTICLE{palensky2011,
  author  = {P. Palensky and D. Dietrich},
  title   = {Demand side management: {Demand} response, intelligent energy systems, and smart loads},
  journal = {IEEE Trans. Ind. Informat.},
  volume  = {7},
  number  = {3},
  pages   = {381--388},
  month   = aug,
  year    = {2011},
  doi     = {10.1109/TII.2011.2158841},
  note    = {doi: 10.1109/TII.2011.2158841}
}

@ARTICLE{saad2012,
  author  = {W. Saad and Z. Han and H. V. Poor and T. Ba{\c{s}}ar},
  title   = {Game-theoretic methods for the smart grid: {An} overview of microgrid systems, demand-side management, and smart grid communications},
  journal = {IEEE Signal Process. Mag.},
  volume  = {29},
  number  = {5},
  pages   = {86--105},
  month   = sep,
  year    = {2012},
  doi     = {10.1109/MSP.2012.2186410},
  note    = {doi: 10.1109/MSP.2012.2186410}
}

@ARTICLE{radosavljevic2016,
  author  = {J. Radosavljevi{\'c} and M. Jevti{\'c} and D. Klimenta},
  title   = {Energy and operation management of a microgrid using particle swarm optimization},
  journal = {Eng. Optim.},
  volume  = {48},
  number  = {5},
  pages   = {811--830},
  year    = {2016},
  doi     = {10.1080/0305215X.2015.1057135},
  note    = {doi: 10.1080/0305215X.2015.1057135}
}

@ARTICLE{du2020,
  author  = {Y. Du and F. Li},
  title   = {Intelligent multi-microgrid energy management based on deep neural network and model-free reinforcement learning},
  journal = {IEEE Trans. Smart Grid},
  volume  = {11},
  number  = {2},
  pages   = {1066--1076},
  month   = mar,
  year    = {2020},
  doi     = {10.1109/TSG.2019.2930299},
  note    = {doi: 10.1109/TSG.2019.2930299}
}

@ARTICLE{ajagekar2019,
  author  = {A. Ajagekar and F. You},
  title   = {Quantum computing for energy systems optimization: {Challenges} and opportunities},
  journal = {Energy},
  volume  = {179},
  pages   = {76--89},
  month   = jul,
  year    = {2019},
  doi     = {10.1016/j.energy.2019.04.186},
  note    = {doi: 10.1016/j.energy.2019.04.186}
}

@ARTICLE{eskandarpour2020,
  author  = {R. Eskandarpour and K. Ghosh and A. Khodaei and A. Paaso and L. Zhang},
  title   = {Quantum-enhanced grid of the future: {A} primer},
  journal = {IEEE Access},
  volume  = {8},
  pages   = {188993--189002},
  year    = {2020},
  doi     = {10.1109/ACCESS.2020.3031595},
  note    = {doi: 10.1109/ACCESS.2020.3031595}
}

@ARTICLE{morstyn2023,
  author  = {T. Morstyn},
  title   = {Annealing-based quantum computing for combinatorial optimal power flow},
  journal = {IEEE Trans. Smart Grid},
  volume  = {14},
  number  = {2},
  pages   = {1093--1102},
  month   = mar,
  year    = {2023},
  doi     = {10.1109/TSG.2022.3200590},
  note    = {doi: 10.1109/TSG.2022.3200590}
}

@ARTICLE{nikmehr2022,
  author  = {N. Nikmehr and P. Zhang and M. A. Bragin},
  title   = {Quantum distributed unit commitment: {An} application in microgrids},
  journal = {IEEE Trans. Power Syst.},
  volume  = {37},
  number  = {5},
  pages   = {3592--3603},
  month   = sep,
  year    = {2022},
  doi     = {10.1109/TPWRS.2022.3141794},
  note    = {doi: 10.1109/TPWRS.2022.3141794}
}

@INPROCEEDINGS{koretsky2021,
  author    = {S. Koretsky and P. Gokhale and J. M. Baker and J. Viszlai and H. Zheng and N. Gurung and R. Burg and E. A. Paaso and A. Khodaei and R. Eskandarpour and F. T. Chong},
  title     = {Adapting quantum approximation optimization algorithm ({QAOA}) for unit commitment},
  booktitle = {Proc. IEEE Int. Conf. Quantum Comput. Eng. (QCE)},
  pages     = {181--187},
  year      = {2021},
  doi       = {10.1109/QCE52317.2021.00035},
  note    = {doi: 10.1109/QCE52317.2021.00035}
}

@MISC{farhi2014,
  author        = {E. Farhi and J. Goldstone and S. Gutmann},
  title         = {A quantum approximate optimization algorithm},
  year          = {2014},
  eprint        = {1411.4028},
  archivePrefix = {arXiv},
  primaryClass  = {quant-ph},
  note          = {arXiv:1411.4028}
}

@ARTICLE{lucas2014,
  author  = {A. Lucas},
  title   = {Ising formulations of many {NP} problems},
  journal = {Front. Phys.},
  volume  = {2},
  pages   = {5},
  month   = feb,
  year    = {2014},
  doi     = {10.3389/fphy.2014.00005},
  note    = {doi: 10.3389/fphy.2014.00005}
}

@ARTICLE{glover2019,
  author  = {F. Glover and G. Kochenberger and Y. Du},
  title   = {Quantum bridge analytics {I}: {A} tutorial on formulating and using {QUBO} models},
  journal = {4OR},
  volume  = {17},
  number  = {4},
  pages   = {335--371},
  month   = dec,
  year    = {2019},
  doi     = {10.1007/s10288-019-00424-y},
  note    = {doi: 10.1007/s10288-019-00424-y}
}

@ARTICLE{preskill2018,
  author  = {J. Preskill},
  title   = {Quantum computing in the {NISQ} era and beyond},
  journal = {Quantum},
  volume  = {2},
  pages   = {79},
  month   = aug,
  year    = {2018},
  doi     = {10.22331/q-2018-08-06-79},
  note    = {doi: 10.22331/q-2018-08-06-79}
}

@ARTICLE{zhou2020,
  author  = {L. Zhou and S.-T. Wang and S. Choi and H. Pichler and M. D. Lukin},
  title   = {Quantum approximate optimization algorithm: {Performance}, mechanism, and implementation on near-term devices},
  journal = {Phys. Rev. X},
  volume  = {10},
  number  = {2},
  pages   = {021067},
  month   = jun,
  year    = {2020},
  doi     = {10.1103/PhysRevX.10.021067},
  note    = {doi: 10.1103/PhysRevX.10.021067}
}

@ARTICLE{cerezo2021,
  author  = {M. Cerezo and A. Arrasmith and R. Babbush and S. C. Benjamin and S. Endo and K. Fujii and J. R. McClean and K. Mitarai and X. Yuan and L. Cincio and P. J. Coles},
  title   = {Variational quantum algorithms},
  journal = {Nat. Rev. Phys.},
  volume  = {3},
  number  = {9},
  pages   = {625--644},
  month   = sep,
  year    = {2021},
  doi     = {10.1038/s42254-021-00348-9},
  note    = {doi: 10.1038/s42254-021-00348-9}
}

@BOOK{bratman1987,
  author    = {M. E. Bratman},
  title     = {Intention, Plans, and Practical Reason},
  publisher = {Harvard Univ. Press},
  address   = {Cambridge, MA, USA},
  year      = {1987}
}

@INPROCEEDINGS{rao1995,
  author    = {A. S. Rao and M. P. Georgeff},
  title     = {{BDI} agents: {From} theory to practice},
  booktitle = {Proc. 1st Int. Conf. Multi-Agent Syst. (ICMAS)},
  address   = {San Francisco, CA, USA},
  pages     = {312--319},
  year      = {1995}
}

@ARTICLE{ioannou2020,
  author  = {I. Ioannou and V. Vassiliou and C. Christophorou and A. Pitsillides},
  title   = {Distributed artificial intelligence solution for {D2D} communication in {5G} networks},
  journal = {IEEE Syst. J.},
  volume  = {14},
  number  = {3},
  pages   = {4232--4241},
  month   = sep,
  year    = {2020},
  doi     = {10.1109/JSYST.2020.2979044},
  note    = {doi: 10.1109/JSYST.2020.2979044}
}

@ARTICLE{ioannou2022cn,
  author  = {I. Ioannou and C. Christophorou and V. Vassiliou and A. Pitsillides},
  title   = {A novel distributed {AI} framework with {ML} for {D2D} communication in {5G/6G} networks},
  journal = {Comput. Netw.},
  volume  = {211},
  pages   = {108987},
  month   = jul,
  year    = {2022},
  doi     = {10.1016/j.comnet.2022.108987},
  note    = {doi: 10.1016/j.comnet.2022.108987}
}

@ARTICLE{ioannou2022access,
  author  = {I. I. Ioannou and C. Christophorou and V. Vassiliou and M. Lestas and A. Pitsillides},
  title   = {Dynamic {D2D} communication in {5G/6G} using a distributed {AI} framework},
  journal = {IEEE Access},
  volume  = {10},
  pages   = {62772--62799},
  year    = {2022},
  doi     = {10.1109/ACCESS.2022.3182388},
  note    = {doi: 10.1109/ACCESS.2022.3182388}
}

@INPROCEEDINGS{ioannou2025planlib,
  author    = {I. Ioannou and A. Gregoriades and P. Nagaradjane and C. Christophorou and V. Vassiliou},
  title     = {An accurate intelligent plan library for belief-based desire prioritization to intentions in {BDIx} agents},
  booktitle = {Proc. Int. Conf. Wireless Commun. Signal Process. Netw. (WiSPNET)},
  pages     = {1--6},
  year      = {2025},
  doi       = {10.1109/WiSPNET64060.2025.11005173},
  note    = {doi: 10.1109/WiSPNET64060.2025.11005173}
}

@BOOK{daibook,
  author    = {I. Ioannou and P. Nagaradjane and V. Vassiliou and A. Pitsillides and C. Christophorou},
  title     = {Distributed Artificial Intelligence for {5G/6G} Communications: Frameworks with Machine Learning},
  publisher = {CRC Press},
  address   = {Boca Raton, FL, USA},
  year      = {2024},
  doi       = {10.1201/9781003469209},
  note    = {doi: 10.1201/9781003469209}
}

@ARTICLE{ioannou2024access,
  author  = {I. I. Ioannou and S. Javaid and C. Christophorou and V. Vassiliou and A. Pitsillides and Y. Tan},
  title   = {A distributed {AI} framework for nano-grid power management and control},
  journal = {IEEE Access},
  volume  = {12},
  pages   = {43350--43377},
  year    = {2024},
  doi     = {10.1109/ACCESS.2024.3377926},
  note    = {doi: 10.1109/ACCESS.2024.3377926}
}

@ARTICLE{glover1990,
  author  = {F. Glover},
  title   = {Tabu search, {Part II}},
  journal = {ORSA J. Comput.},
  volume  = {2},
  number  = {1},
  pages   = {4--32},
  year    = {1990},
  doi     = {10.1287/ijoc.2.1.4},
  note    = {doi: 10.1287/ijoc.2.1.4}
}

@ARTICLE{kirkpatrick1983,
  author  = {S. Kirkpatrick and C. D. Gelatt and M. P. Vecchi},
  title   = {Optimization by simulated annealing},
  journal = {Science},
  volume  = {220},
  number  = {4598},
  pages   = {671--680},
  month   = may,
  year    = {1983},
  doi     = {10.1126/science.220.4598.671},
  note    = {doi: 10.1126/science.220.4598.671}
}

@INPROCEEDINGS{kennedy1997,
  author    = {J. Kennedy and R. C. Eberhart},
  title     = {A discrete binary version of the particle swarm algorithm},
  booktitle = {Proc. IEEE Int. Conf. Syst., Man, Cybern.},
  address   = {Orlando, FL, USA},
  volume    = {5},
  pages     = {4104--4108},
  year      = {1997},
  doi       = {10.1109/ICSMC.1997.637339},
  note    = {doi: 10.1109/ICSMC.1997.637339}
}

@BOOK{liu2009coop,
author    = {K. J. Ray Liu and Ahmed K. Sadek and Weifeng Su and Andres Kwasinski},
title     = {Cooperative Communications and Networking},
publisher = {Cambridge University Press},
address   = {Cambridge, U.K.},
year      = {2009},
doi       = {10.1017/CBO9780511754524},
  note    = {doi: 10.1017/CBO9780511754524}
}

\end{document}